%% file: main.tex
\documentclass[10pt,conference,compsocconf]{IEEEtran}

\usepackage{hyperref}
\usepackage{graphicx}	
\usepackage{amsmath}
\usepackage{caption}
\usepackage[noabbrev]{cleveref}
\usepackage{svg}

\usepackage{tikz}

\newcommand*\circled[1]{\tikz[baseline=(char.base)]{\node[shape=circle,draw,inner sep=1.3pt] (char) {#1};}}
\newcommand*\smallcircled[1]{\tikz[baseline=(char.base)]{\node[shape=circle,draw,inner sep=0.05pt] (char) {#1};}}

\usepackage{enumitem}

\title{Packed-Ensemble Surrogate Models for Fluid Flow Estimation Arround Airfoil Geometries}

\author{
  Anthony Kalaydjian -- Anton Balykov -- Alexi Semiz\\
  \textit{Department of Computer Science, EPFL, Switzerland}\\
\vspace{-8pt}
  \\
  Under the supervision of Adrien Chan-Hon-Tong\\
  \textit{Onera, Université Paris Saclay, France}
}

\begin{document}
\thispagestyle{plain}

\maketitle
\input{abstract}

\input{introduction}

\input{the_data}

\input{packed_ensemble}

\input{our_method}

\input{conclusion}

\newpage
\input{ethics}

\newpage
\bibliographystyle{unsrt}  
\bibliography{references}   

\input{appendix}

\end{document}

%% file: abstract.tex
\begin{abstract}
Physical based simulations can be very time and computationally demanding tasks. One way of accelerating these processes is by making use of data-driven surrogate models that learn from existing simulations.
Ensembling methods are particularly relevant in this domain as their smoothness properties coincide with the smoothness of physical phenomena. 
The drawback is that they can remain costly. This research project focused on studying Packed-Ensembles that generalize Deep Ensembles but remain faster to train. Several models have been trained and compared in terms of multiple important metrics. PE(8,4,1) has been identified as the clear winner in this particular task, beating down its Deep Ensemble conterpart while accelerating the training time by 25\%.
\end{abstract}

%% file: introduction.tex
\section{Introduction}

Reliance on simulations is crucial in industry to optimize a physical system's properties, yet existing simulators face huge limitations in computation time and resource when dealing with complex models. 
Augmenting simulations with data-driven approaches shows promise but often faces an accuracy-speed trade-off.

The problem that is tackled here relates to the optimization of an airfoil design, and is part of the \href{https://www.codabench.org/competitions/1534}{ML4PhySim} challenge hosted by IRT-SystemX. More precisely, the goal is to accelerate the estimation (compared to classical simulations) of the flow quantities around an airfoil geometry by making use of data-driven surrogate models.

This project also tackles an ML interest point in processing of physical inputs, as these types of data follow physical laws in contrast with most of the data that is considered by ML applications that doesn't follow any particular structure (pictures, etc.).

Ensembling methods such as Deep Ensembles are very relevant for such physics problems, as they tend to smooth out the results of classical models, which better fits to the smoothness of physical phenomena.  
This study will focus on Packed-Ensembles (PE) \cite{packedensembles} which generalize the idea of ensembles by embedding several smaller models in a lightweight architecture. Compared to Deep Ensembles, this method allows to train in parallel several small models, independently, as part of a big model. Averaging the output of the small models allows to reduce the uncertainty of the output, while reducing the cost of using a classical, larger, Deep-Ensemble which operates multiple distinct models.

The data that will be used is the AirfRANS dataset \cite{airfrans}, which consists of several Computational Fluid Dynamics (CFD) simulations on different airfoil geometries.

The study presented here will be focused on two main axis. The first axis is the exploration of the PE's parameters on the given problem, for which a cross validation has been implemented. The second axis will focus on the exploitation of the "Learning Industrial Physical Simulations" (LIPS) platform \cite{LIPS}, a versatile framework that allows to compute relevant metrics for evaluating the quality of the proposed model.

\begin{figure}[!h]
\includegraphics[width=7cm]{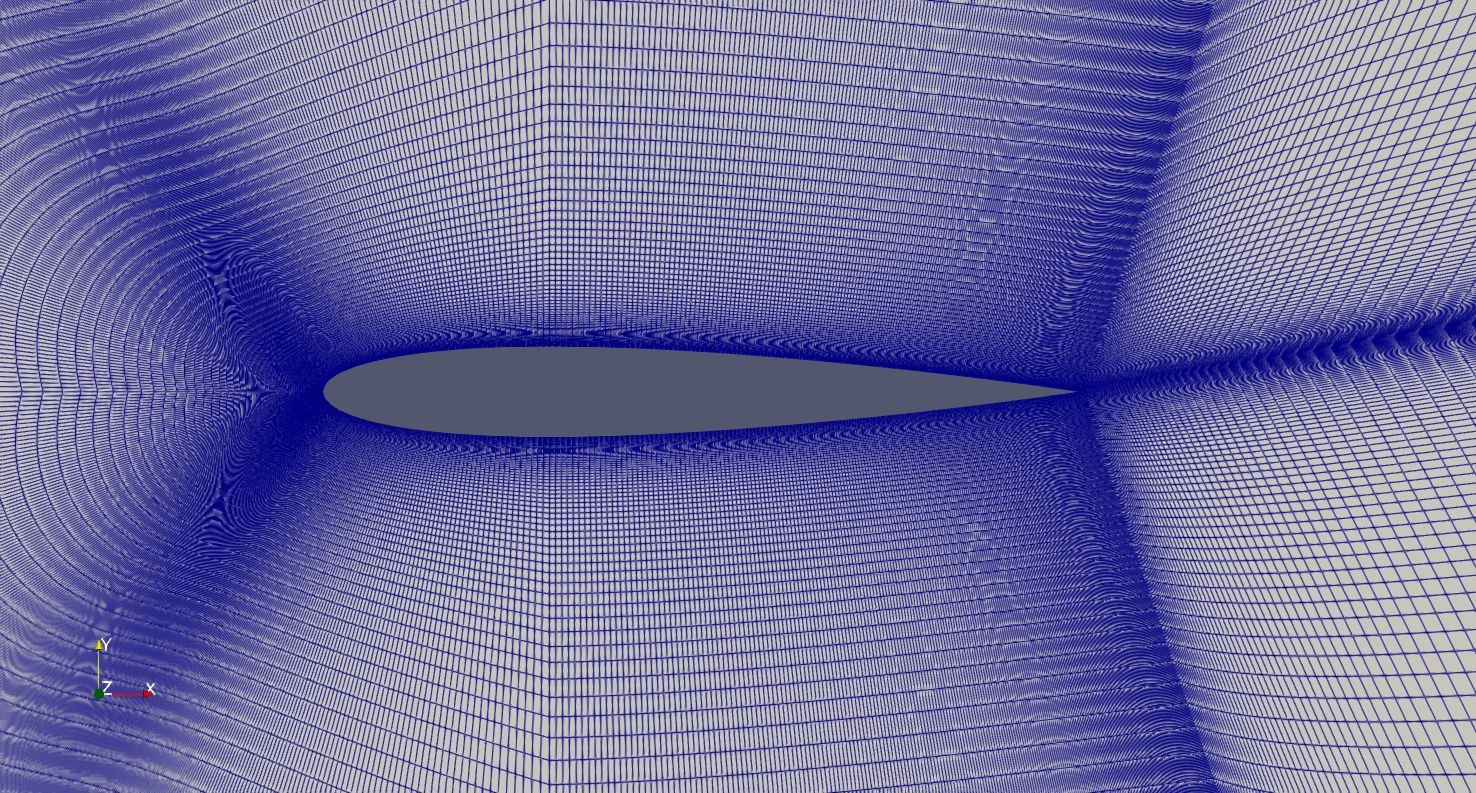}
\centering
\caption{Training Mesh -- \href{https://www.codabench.org/competitions/1534}{ML4PhySim}}
\label{fig:mesh}
\end{figure}

%% file: the_data.tex
\section{Material \& Methods}

\subsection{The data}

The AirfRANS dataset that is used contains several CFD simulations of the incompressible Reynolds-Averaged Navier–Stokes (RANS) equations in a subsonic flight regime setup, which are themselves approximations of the full-on Navier-Stokes equations. The challenge that is proposed is thus the first step towards building surrogate models for the complete Direct Numerical Simulation (DNS) problem.

The provided CFD Simulations are computed over several airfoil geometries that were parametrized using the NACA 4 and 5 digits series of airfoils \cite{NACA}. Each simulation consists of point-clouds on a mesh around the airfoil as seen in \cref{fig:mesh}, with each point being described via 7 features:

\begin{itemize}
    \item \textbf{position} : $x$ and $y$ positions in m of the point on the mesh.
    \item \textbf{inlet velocity} : inlet velocities at the point in m/s.
    \item \textbf{distance to the airfoil} : distance in m.
    \item \textbf{normals} : two normal components to the airfoil in m, set to 0 if the point is not on the airfoil.
\end{itemize}

Each point is also given a target of 4 components for the underlying regression task:

\begin{itemize}
    \item \textbf{velocity}                         : $x$ and $y$ velocities of the fluid at that point in m/s.
    \item \textbf{pressure}                         : pressure divided by the specific mass in m²/s².
    \item \textbf{turbulent kinematic viscosity}    : one component in m²/s.
\end{itemize}

The provided dataset contains three subsets. A training set with $103$ simulations, a regular test-set with $200$ simulations used to evaluate the interpolation capabilities of the trained model, and an OOD test-set of size $496$ used to evaluate the extrapolation capabilities of the model.

The model will thus be defined with an input-size of $7$ and an output size of $4$ and will be evaluated sequentially on each point of each considered simulation.

The average number of nodes of a simulation per set, as well as the total number of points per set are shown in \cref{tab:avg_num_pts}.

\begin{table}[!h]
    \centering
    \begin{tabular}{|c||c|c|c|}
        \hline
         Dataset                & training   & test         & test-OOD\\ \hline \hline
         avg number of nodes    & 179761     & 179246       & 179586\\ \hline
         Dataset size           & 18515415   & 35849332     & 89074648\\
         \hline
    \end{tabular}
    \caption{Average number of mesh-points per simulation per set \& total number of points per set}
    \label{tab:avg_num_pts}
\end{table}

\subsection{Data pre-processing}

The provided data was generated with the physics simulator which provides the full mesh for each simulation. For the current task all the data is numerical and continuous and we do not have any missing data.

The input features are standardized (by fitting a standard scaler to the training data) as to improve the distribution of the input data for training. The output features are also scaled.

The original scale of the prediction is then retrieved by applying the inverse transformation to the output of our model.

%% file: packed_ensemble.tex
\subsection{Packed-Ensembles}

Given that the dynamical nature of the system the data is coming from is complex and may vary largely from one airfoil to another, we may need to use complex models to fit the data. The problem with these models is that they have a large variance and their result might not be as smooth as they should.

\begin{figure}[!h]
    \centering
    \includegraphics[width=0.25\textwidth]{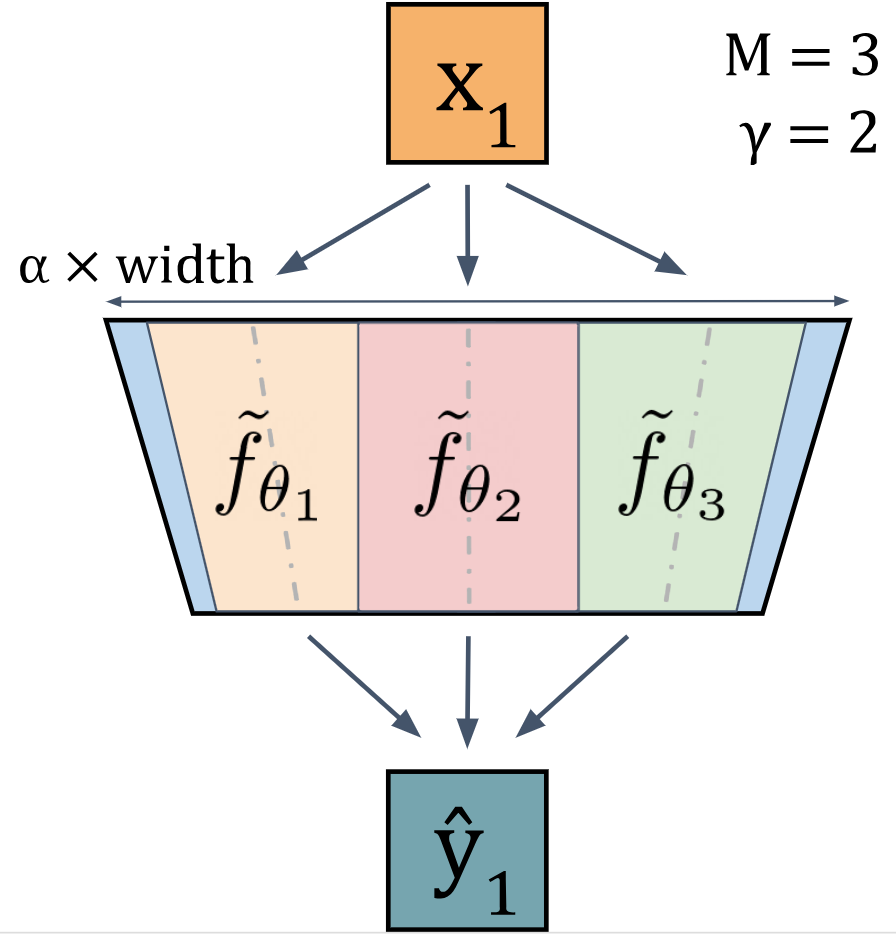}
    \caption{Packed ensemble architecture for $\alpha$, $M = 3$, $\gamma = 2$} \cite{packedensembles}
    \label{fig:PE_figure}
\end{figure}

Deep Ensembles (DEs) \cite{deepensembles} have been used to solve this problem. The idea behind this approach is to train several independent models and creating an estimator that averages the outputs of all models.
Although DEs provide undisputed benefits, they also come with the significant drawback that the training time and the memory usage in inference increases linearly with the number of networks.
The Packed-Ensembles architecture has thus been introduced to alleviate these problems. It deals with these by embedding small sub-networks, which are essentially DNNs with fewer parameters, into the original network, as seen in \cref{fig:PE_figure}.
The torch-uncertainty package provides Packed convolutional layers as well as linear layers that will be used to create and train our models.

By employing grouped convolutions, Packed-Ensembles partition each layer, creating distinct sub-networks operating in parallel. These sub-networks enable simultaneous computations, enhancing efficiency without the computational burden of training entirely separate models. PEs are not just parallel implementations of DEs, but come with a set of hyperparameters ($M$, $\alpha$, $\gamma$) that allow to tune aspects of the sub-networks.

The number of estimators $M$ defines the number of separate sub-models that would train in parallel.

Since the PE approach embeds smaller sub-models that have a fraction of the number of parameters of the initial model architecture, they have (taken separately) a lower representational capability. The capacity modulator $\alpha$ fixes this and allows to multiplicatively adjust the number of parameters of the sub-networks to strike a balance between reduced number of trained parameters and ensemble performance.

On the other hand, the sparsity modulator $\gamma$, allows to reduce the number of weights within sub-networks, which could ideally result in  having a globally smaller network that perform not much worse.

%% file: our_method.tex
\section{Experimental protocol \& Results}

\subsection{Hyperparameters space}

As it was mentioned previously, we have chosen a Packed-Ensemble Multilayer Perceptron (MLP) as our model architecture due to its ensemble performance for reduced cost. To select the best model for the current task, we first need to tune the hyperparameters of the model. As in general Deep Learning (DL) methods, we first have:

\begin{itemize}
    \item \textbf{learning rate} of the optimizer
    \item \textbf{dropout} layers existence
\end{itemize}

To these, we add the PE hyperparameters, namely $M$, $\alpha$ and $\gamma$ that were previously mentioned.

\subsection{Cross-validation exploration}

It is interesting, to get familiar with this new model, to experiment the influence of the different hyperparameters of it. To do so, we performed a Cross Validation a grid of different hyperparameter values.
Due to the huge size of the train dataset, we decided to only use a third of it. We used, for this process, the following architecture: input layer (7 $\rightarrow$ 48), hidden layer (48 $\rightarrow$ 128), hidden layer (128 $\rightarrow$ 48) and output layer (48 $\rightarrow$ 4) and the ReLU as the activation function. A dropout layer with probability 0.2 could also be added after each activation layer, and is controlled by the \textit{droput} boolean. The number of estimators $M$ was chosen constant equal to 4, as the most interesting behavior is for the other parameters. Each model was trained using a 4-fold cross-validation leaving 1 fold for the validation and training on the remaining 3. The performance of each model was calculated as the validation loss, averaged over all 4 iterations for different left out folds.
Each model was trained for 100 epochs maximum, but stopped the training process if the change in the loss was lower than 1\% comparing with the previous epoch for five epochs.

The results for different values on the hyperparameters space are shown in the \cref{tab:cv_results}. We can draw multiple highly valuable conclusions from this process. First of all, the model without dropout layers, with $\alpha=4$, $\gamma=2$ and the lower learning rate had poor performance (0.757). This may be due to the small fraction of data, clearly not enough to train so much more parameters comparing to the other models. It is clear from the table that models which were trained with lower learning rate had worse performance. Again, this is probably due to the small fraction of data used.
On the other hand, for this setting the best performance (0.179) was shown by the model without dropout layers, with $\alpha=2$, $\gamma=2$ and higher learning rate of 0.01. Interestingly, almost the same performance (0.195) could be achieved with dropout layers, $\alpha$=4, $\gamma$=4 and learning rate of 0.01. This means that dropout indeed makes sense for the models with more parameters (the same setting without dropout layers gave us a validation loss of 0.293).

One other finding is the fact that the performance of the model with $\alpha=2$, $\gamma=2$ and a learning rate of $0.01$ is very close to the same model with $\alpha=4$. This means that in this setting, we were able to drastically cut the number of parameters of our estimator compared to a deep ensemble ($\alpha=M$) while staying close to its performance.

These findings will be used in the future steps of our model selection process.

\begin{table}
    \centering
    \begin{tabular}{|l|c|c|l||c|}
    \hline
        dropout & alpha & gamma & learning rate & validation\_loss \\ \hline\hline
        True & 2 & 2 & 0.01 & 0.205 \\ \hline
        True & 2 & 2 & 0.001 & 0.538 \\ \hline
        True & 2 & 4 & 0.01 & 0.261 \\ \hline
        True & 2 & 4 & 0.001 & 0.757 \\ \hline
        True & 4 & 2 & 0.01 & 0.176 \\ \hline
        True & 4 & 2 & 0.001 & 0.243 \\ \hline
        True & 4 & 4 & 0.01 & 0.195 \\ \hline
        True & 4 & 4 & 0.001 & 0.437 \\ \hline
        False & 2 & 2 & 0.01 & 0.179 \\ \hline
        False & 2 & 2 & 0.001 & 0.351 \\ \hline
        False & 4 & 4 & 0.01 & 0.293 \\ \hline
        False & 4 & 4 & 0.001 & 0.456 \\ \hline
    \end{tabular}
    \caption{Cross-validation results}
    \label{tab:cv_results}
\end{table}

\subsection{LIPS}

Although we train our model on a Machine Learning (ML) loss, it is very interesting, in this physical setting, to consider other more meaningful metrics. 
One way to do so is to use the LIPS platform, which evaluates our model on a list of metrics among which, the ones that we mainly focused on:

\begin{itemize}
    \item \textbf{MSE} of:
    \begin{itemize}
        \item \textbf{x-velocity} : the lower the better
        \item \textbf{y-velocity} : the lower the better
        \item \textbf{pressure} : the lower the better
        \item \textbf{surface pressure} : the lower the better
        \item \textbf{turbulent viscosity} : the lower the better
    \end{itemize}
    \item \textbf{mean relative drag} : the lower the better
    \item \textbf{mean relative lift} : the lower the better
    \item \textbf{Spearman's correlation for drag} : the higher the better
    \item \textbf{Spearman's correlation for lift} : the higher the better
\end{itemize}

While the first 5 metrics of the list are minimized directly by the model (surface pressure is calculated afterwards), the last 4 are of much higher interest for the current setting as they show how well the relationship between our model estimate and the ground-truth are related.

The Spearman's correlations are of particular interest as minimizing the true lift over drag ratio could be done by minimizing its estimates if the relationships between our estimates and the ground truth can be well explained by an increasing function (which is the case when that coefficient is close to 1).

\subsection{Model selection}
\label{sec:model_selection}

The Packed-Ensemble MLP was then adapted in accordance to the requirements of the LIPS platform benchmarking.
Each model have then been trained for 200 epochs on the same training dataset. The number of estimator of each model has been fixed $M=8$ to be able to cover a wider range of values for $\alpha$ and $\gamma$.
Among the various hyperparameters we chose to test, we find, \textbf{layers}, \textbf{$\alpha$}, \textbf{$\gamma$}, \textbf{dropout layers}, \textbf{learning rate} (lr) of the Adam optimizer and the \textbf{weight decay} of the Adam optimizer.

Multiple configurations were tried as shown in the \cref{tab:models}.

\begin{table*}
\centering
\begin{tabular}{|c||l|c|c|c|l|c|l||l|l|l|l|} 
\cline{2-12}
\multicolumn{1}{l|}{} & \multicolumn{7}{c||}{Packed-MLP hyperparameters}                                                                                      & \multicolumn{4}{c|}{Test results}                                                                                                                                                                                                                                                    \\ 
\hline
model                 & layers                              & $M$        & $\alpha$   & $\gamma$   & dropout        & lr            & weight decay  & \begin{tabular}[c]{@{}l@{}}mean\\relative\\drag\end{tabular} & \begin{tabular}[c]{@{}l@{}}mean\\relative\\lift\end{tabular} & \begin{tabular}[c]{@{}l@{}}Spearman's\\correlation\\for drag\end{tabular} & \begin{tabular}[c]{@{}l@{}}Spearman's\\correlation\\for lift\end{tabular}  \\ 
\hline \hline
\circled{1}                     & (48,128,48)                         & 4          & 2          & 2          & False          & 1e-2          & False         & 3.775                                                        & 1.587                                                        & 0.105                                                                     & 0.052                                                                      \\ 
\hline
\circled{2}                     & (48,128,256,128,48)                 & 8          & 4          & 2          & False          & 1e-2          & False         & 2.221                                                        & 1.271                                                        & 0.025                                                                     & 0.020                                                                       \\ 
\hline
\circled{3}                     & (64,64,8,64,64,64,8,64,64)          & 8          & 4          & 1          & False          & 2e-4          & False         & 26.256                                                       & 0.643                                                        & 0.308                                                                     & 0.937                                                                      \\ 
\hline
\circled{4}                     & (64,32,16,32,64,32,16,32,64)        & 8          & 4          & 1          & False          & 2e-4          & False         & 26.257                                                       & 0.643                                                        & 0.308                                                                     & 0.937                                                                      \\ 
\hline
\circled{5}                     & (64,64,64,8,64,64,64,64,8,64,64,64) & 8          & 4          & 1          & False          & 2e-4          & False         & 26.894                                                       & 0.532                                                        & 0.314                                                                     & 0.948                                                                      \\ 
\hline
\circled{6}                     & (64,64,64,64,64,64,64,64,64,64)     & 8          & 4          & 1          & False          & 2e-4          & False         & 26.257                                                       & 0.643                                                        & 0.308                                                                     & 0.937                                                                      \\ 
\hline
\circled{7}                     & (32,32,32,32,32,32,32,32,32,32)     & 8          & 4          & 1          & False          & 2e-4          & False         & 26.894                                                       & 0.532                                                        & 0.314                                                                     & 0.948                                                                      \\ 
\hline
\circled{8}                     & (64,64,8,64,64,64,8,64,64)          & 8          & 1          & 1          & False          & 2e-4          & False         & 14.004                                                       & 1.184                                                        & 0.286                                                                     & 0.740                                                                      \\ 
\hline
\circled{9}                     & (64,64,8,64,64,64,8,64,64)          & 8          & 2          & 1          & False          & 2e-4          & False         & 20.476                                                       & 0.622                                                        & 0.196                                                                     & 0.935                                                                      \\ 
\hline
\smallcircled{10}                    & (64,64,8,64,64,64,8,64,64)          & 8          & 6          & 1          & False          & 2e-4          & False         & 28.816                                                      & 0.546                                                        & 0.247                                                                     & 0.911                                                                      \\ 
\hline
\smallcircled{11}                    & (64,64,8,64,64,64,8,64,64)          & 8          & 8          & 1          & False          & 2e-4          & False         & 28.420                                                        & 0.697                                                        & 0.220                                                                      & 0.946                                                                      \\ 
\hline
\smallcircled{12}                    & (64,64,8,64,64,64,8,64,64)          & 8          & 4          & 2          & False          & 2e-4          & 1e-5          & 24.320                                                        & 0.556                                                        & 0.218                                                                     & 0.916                                                                      \\ 
\hline
\smallcircled{13}                    & (64,64,8,64,64,64,8,64,64)          & 8          & 4          & 4          & False          & 2e-4          & 1e-5          & 14.665                                                       & 0.552                                                        & 0.196                                                                     & 0.929                                                                      \\ 
\hline
\smallcircled{14}                    & (64,64,8,64,64,64,8,64,64)          & 8          & 2          & 1          & False          & 2e-4          & 1e-5          & 17.490                                                        & 0.980                                                         & 0.186                                                                     & 0.875                                                                      \\ 
\hline
\smallcircled{\textbf{15}}           & \textbf{(64,64,8,64,64,64,8,64,64)} & \textbf{8} & \textbf{4} & \textbf{1} & \textbf{False} & \textbf{2e-4} & \textbf{1e-5} & \textbf{25.568}                                              & \textbf{0.549}                                               & \textbf{0.310}                                                             & \textbf{0.957}                                                             \\ 
\hline
\smallcircled{16}                    & (64,64,8,64,64,64,8,64,64)          & 8          & 6          & 1          & False          & 2e-4          & 1e-5          & 28.298                                                       & 0.450                                                         & 0.255                                                                     & 0.923                                                                      \\ 
\hline
\smallcircled{17}                    & (64,64,8,64,64,64,8,64,64)          & 8          & 8          & 1          & False          & 2e-4          & 1e-5          & 27.410                                                        & 0.527                                                        & 0.282                                                                     & 0.946                                                                      \\ 
\hline
\smallcircled{18}                    & (64,64,8,64,64,64,8,64,64)          & 8          & 4          & 1          & True           & 2e-4          & 1e-5          & 18.570                                                        & 1.270                                                         & 0.045                                                                     & 0.800                                                                       \\ 
\hline
\smallcircled{19}                    & (64,64,8,64,64,64,8,64,64)          & 8          & 6          & 1          & True           & 2e-4          & 1e-5          & 21.490                                                        & 1.342                                                        & 0.159                                                                     & 0.832                                                                      \\
\hline
\end{tabular}
\caption{List of Packed-Ensemble models and their evaluation results on the test set}
\label{tab:models}
\end{table*}

To begin with, for the given task and the given the Packed-Ensemble approach, a thin architecture of the model is much better than a thick one (models \circled{1} \& \circled{2} in the \cref{tab:models}). The difference of the metrics values isn't very significant among the deep and thin models with slightly different configurations of layers (models \circled{3} to \circled{7} in the \cref{tab:models}).

One observation that can be made is that there is a trade-off between the accuracy of the physical metrics. For instance, a PE model that performs better in one of the physical metrics (e.g. drag), will tend to perform more poorly for another metric (e.g. lift, respectively), as for models \circled{9} \& \smallcircled{10}.

After experimenting with different values of $\alpha$ it is clear that, in general, models with bigger values of $\alpha$ tend to perform better, as for models \circled{8} to \smallcircled{11}. Although it is tempting to take $\alpha$ smaller for a smaller overall model, $\alpha$=1 and $\alpha$=2 resulted in significantly worse results.

Training models with different values for $\gamma$ also shows that adding more sparsity to the model leads to worse results in terms of machine learning metrics, as in models \smallcircled{12} \& \smallcircled{13}. However, these differences for $\gamma=2$ and $\gamma=4$ are not large enough to draw conclusions.

Adding dropout layers (models \smallcircled{18} \& \smallcircled{19} in the \cref{tab:models}) resulted in worse performance, especially in terms of physical metrics. Indeed, adding dropout layers to thin models reduces their representation capabilities.

Learning rate that had previously shown promising results during the cross-validation (\cref{tab:cv_results}), looked too large for this dataset instance. For most of the experiments the learning rate that was used was $2\mathrm{e}{-4}$.

The baseline for this task is a classic MLP with (64,64,8,64,64,64,8,64,64) architecture, trained for 200 epochs with Adam optimizer with the learning rate of $2\mathrm{e}{-4}$. The scores of this model are:

\begin{itemize}
    \item \textbf{MSE}: ("x-velocity": 849.105, "y-velocity": 992.3, "pressure": 7932974.373, "turbulent viscosity and surface pressure": 0.0002)
    \item \textbf{Physics}: ("mean relative drag": 1.006, "mean relative lift": 0.994, "Spearman's correlation for drag": -0.05, "Spearman's correlation for lift": 0.038)
\end{itemize}

Adding regularization via a weight decay of $1\mathrm{e}{-5}$ to the Adam optimizer helps avoiding the possible over-fitting of the model and yields better results (models \smallcircled{14} to \smallcircled{17}).

\subsection{Results}

One interesting thing to notice is that although our PE models are very accurate for the lift components of the physics metrics, they are way less for its drag components, as we notice Spearman's correlations that can only reach up to about 0.3, against 0.96 for lift. This is also seen with a global mean relative drag that is usually one order of magnitude larger than that of lift.

Among all of the PEs that were trained, the PE($8$, $4$, $1$) was the best performing one, with very high Spearman's correlations (see model 15 in the \cref{tab:models}). This model even outperformed the classic deep-ensemble model \smallcircled{17}, while having a 25\% faster training time with 6441s for model \smallcircled{15} against 8664s for model \smallcircled{17} on an RTX 4070 GPU using 6 CPU workers.

It is also worth mentioning that it is the same model that performed best on the test-OOD dataset, as shown in the appendix, \cref{tab:ood}.

%% file: conclusion.tex
\section{Conclusion}

This report shows the relevance of Packed-Ensemble models for the regression of physical quantities of a complex system, e.g. fluid fields around an airfoil geometry.
Several architectures of PE were tested and their behaviour and results were compared in terms of ML metrics, as well as more explicit physics-based metrics.

The development of surrogate models for optimizing an airfoil's shape requires high Spearman's rank correlation coefficients for both the lift and drag, so the optimization of the lift over drag ratio using the surrogate model would closely relate to the optimization of the airfoil's true physical ratio.

Across the many models that were evaluated, the PE(8,4,1) with regularization was the one performing best, even outperforming its PE(8,8,1) Deep-Ensemble model for a 25\% faster training time.

Packed-Ensembles can thus can be used for physics-based regression tasks as more reliable models compared to simple MLPs, while being faster alternatives to Deep-Ensembles.

This project also opens the question about whether ML models are capable of converging towards an understanding of physics, as these types of data are governed by physical laws, in contrast with the inputs that are usually considered.

%% file: ethics.tex
\section{Ethical risks}
We have not identified any ethical risk in this study. The first direct stakeholder are airfoil companies. They are looking for a cheaper and more efficient solution to design their products. Similarly physics labs studying fluid flows are also among the first recipients of the project. The training data comes from direct simulations that stem from universal physics laws. Moreover, no personal/human data is involved here. Thus, the models we computed make no human bias. The only stakeholders that could be harmed by this kind of models are simulation companies. Indeed we aim at substituting their solutions for cheaper ones. Nonetheless, the necessary data to train the models directly comes from simulation data. Consequently, simulations software and simulation companies should work hand in hand. In the worst scenario we can imagine that physics simulation software companies may manipulate the data they provide or simply forbid their use for any solution that could replace theirs.
Besides, the projects aims at building surrogate models that are part of a bigger and strict design process, so the eventual flaws or defects returned by such models would get alleviated by the verification procedure.
Moreover, the goal of designing faster training surrogate models (as PE) that make simulations faster is aligned with current concerns about ecological cost of machine learning. However the training and study still has a cost, especially when performing a large grid search for hyper-parameters tuning. One should always consider this aspect when training models for hours. This could be seen as an ecological hazard if large companies try to train bigger models on huge datasets, without being concerned about this risk.

%% file: appendix.tex

\newpage

\onecolumn
\appendix
\label{appendix}

\begin{table}[h]
\centering  
    \begin{tabular}{|c||l|c|c|c|l|c|l||l|l|l|l|} 
    \cline{2-12}
    \multicolumn{1}{l|}{} & \multicolumn{7}{c||}{Packed-MLP hyperparameters}                                                                                      & \multicolumn{4}{c|}{Test results}                                                                                                                                                                                                                                                    \\ 
    \hline
    model                 & layers                              & $M$        & $\alpha$   & $\gamma$   & dropout        & lr            & weight decay  & \begin{tabular}[c]{@{}l@{}}mean\\relative\\drag\end{tabular} & \begin{tabular}[c]{@{}l@{}}mean\\relative\\lift\end{tabular} & \begin{tabular}[c]{@{}l@{}}Spearman's\\correlation\\for drag\end{tabular} & \begin{tabular}[c]{@{}l@{}}Spearman's\\correlation\\for lift\end{tabular}  \\ 
    \hline \hline
    \circled{1}                     & (48,128,48)                         & 4          & 2          & 2          & False          & 1e-2          & False         & 5.60                                                        & 2.181                                                        & 0.074                                                                     & 0.080                                                                      \\ 
    \hline
    \circled{2}                     & (48,128,256,128,48)                 & 8          & 4          & 2          & False          & 1e-2          & False         & 2.94                                                        & 1.33                                                        & 0.066                                                                     & 0.039                                                                       \\ 
    \hline
    \circled{3}                     & (64,64,8,64,64,64,8,64,64)          & 8          & 4          & 1          & False          & 2e-4          & False         & 30.1                                                       & 1.09                                                        & 0.278                                                                     & 0.928                                                                      \\ 
    \hline
    \circled{4}                     & (64,32,16,32,64,32,16,32,64)        & 8          & 4          & 1          & False          & 2e-4          & False         & 30.1                                                       & 1.09                                                        & 0.278                                                                     & 0.928                                                                      \\ 
    \hline
    \circled{5}                     & (64,64,64,8,64,64,64,64,8,64,64,64) & 8          & 4          & 1          & False          & 2e-4          & False         & 30.7                                                       & 0.782                                                        & 0.281                                                                     & 0.939                                                                      \\ 
    \hline
    \circled{6}                     & (64,64,64,64,64,64,64,64,64,64)     & 8          & 4          & 1          & False          & 2e-4          & False         & 30.1                                                       & 1.09                                                        & 0.278                                                                     & 0.928                                                                      \\ 
    \hline
    \circled{7}                     & (32,32,32,32,32,32,32,32,32,32)     & 8          & 4          & 1          & False          & 2e-4          & False         & 30.7                                                       & 0.782                                                        & 0.281                                                                     & 0.939                                                                      \\ 
    \hline
    \circled{8}                     & (64,64,8,64,64,64,8,64,64)          & 8          & 1          & 1          & False          & 2e-4          & False         & 15.8                                                       & 1.71                                                        & 0.237                                                                     & 0.647                                                                       \\ 
    \hline
    \circled{9}                     & (64,64,8,64,64,64,8,64,64)          & 8          & 2          & 1          & False          & 2e-4          & False         &  22.5                                                      & 0.549                                                        & 0.177                                                                     & 0.940                                                                    \\ 
    \hline
    \smallcircled{10}                    & (64,64,8,64,64,64,8,64,64)          & 8          & 6          & 1          & False          & 2e-4          & False         & 33.2                                                      & 0.794                                                        & 0.275                                                                     & 0.896                                                                      \\ 
    \hline
    \smallcircled{11}                    & (64,64,8,64,64,64,8,64,64)          & 8          & 8          & 1          & False          & 2e-4          & False         & 31.8                                                        & 0.736                                                        & 0.244                                                                      & 0.938                                                                      \\ 
    \hline
    \smallcircled{12}                    & (64,64,8,64,64,64,8,64,64)          & 8          & 4          & 2          & False          & 2e-4          & 1e-5          & 27.1                                                        & 0.527                                                        & 0.193                                                                     & 0.948                                                                      \\ 
    \hline
    \smallcircled{13}                    & (64,64,8,64,64,64,8,64,64)          & 8          & 4          & 4          & False          & 2e-4          & 1e-5          & 16.0                                                       & 0.592                                                        & 0.142                                                                     & 0.938                                                                      \\ 
    \hline
    \smallcircled{14}                    & (64,64,8,64,64,64,8,64,64)          & 8          & 2          & 1          & False          & 2e-4          & 1e-5          & 18.8                                                        & 1.64                                                         & 0.143                                                                     & 0.832                                                                      \\ 
    \hline
    \smallcircled{\textbf{15}}           & \textbf{(64,64,8,64,64,64,8,64,64)} & \textbf{8} & \textbf{4} & \textbf{1} & \textbf{False} & \textbf{2e-4} & \textbf{1e-5} & \textbf{29.8}                                              & \textbf{0.83}                                               & \textbf{0.280}                                                             & \textbf{0.961}                                                             \\ 
    \hline
    \smallcircled{16}                    & (64,64,8,64,64,64,8,64,64)          & 8          & 6          & 1          & False          & 2e-4          & 1e-5          & 33.3                                                       & 0.685                                                         & 0.287                                                                     & 0.916                                                                      \\ 
    \hline
    \smallcircled{17}                    & (64,64,8,64,64,64,8,64,64)          & 8          & 8          & 1          & False          & 2e-4          & 1e-5          & 31.5                                                        & 0.503                                                        & 0.319                                                                     & 0.962                                                                      \\ 
    \hline
    \smallcircled{18}                    & (64,64,8,64,64,64,8,64,64)          & 8          & 4          & 1          & True           & 2e-4          & 1e-5          & 23.3                                                        & 1.76                                                         & 0.076                                                                     & 0.702                                                                        \\ 
    \hline
    \smallcircled{19}                    & (64,64,8,64,64,64,8,64,64)          & 8          & 6          & 1          & True           & 2e-4          & 1e-5          & 27.0                                                        & 1.86                                                       & 0.170                                                                     & 0.743                                                                      \\
    \hline
    \end{tabular}

    \label{tab:ood}

    \centering

    \caption{List of Packed-Ensemble models and their evaluation results on the test-OOD set}
    \label{tab:ood}
        
\end{table}